\documentclass[10pt,twocolumn,letterpaper]{article}

\usepackage[pagenumbers]{cvpr} 

\usepackage{graphicx}
\usepackage{amsmath}
\usepackage{amssymb}
\usepackage{booktabs}
\usepackage{multirow}
\usepackage{booktabs, multirow} 
\usepackage{soul}
\usepackage[table]{xcolor}
\usepackage{changepage,threeparttable}
\usepackage[accsupp]{axessibility}
\usepackage{mmstyle}

\definecolor{MyGreen}{HTML}{1FBC03}
\newcommand{\improve}[1]{{\textcolor{MyGreen}{$^{+#1}$}}}
\newcommand{\tablestyle}[2]{\setlength{\tabcolsep}{#1}\renewcommand{\arraystretch}{#2}\centering\small}
\newcommand{\blfootnote}[1]{%
  \begingroup
  \renewcommand\thefootnote{}%
  \footnote{#1}%
  \addtocounter{footnote}{-1}%
  \endgroup
}

\usepackage[pagebackref,breaklinks,colorlinks]{hyperref}
\usepackage[misc]{ifsym}

\usepackage[capitalize]{cleveref}
\crefname{section}{Sec.}{Secs.}
\Crefname{section}{Section}{Sections}
\Crefname{table}{Table}{Tables}
\crefname{table}{Tab.}{Tabs.}

\def\cvprPaperID{5441}
\def\confName{CVPR}
\def\confYear{2023}

\begin{document}

\title{MV-JAR: Masked Voxel Jigsaw and Reconstruction for LiDAR-Based Self-Supervised Pre-Training}

\author{Runsen Xu$^{1,2}$\quad
Tai Wang$^{1,2}$\quad
Wenwei Zhang$^{3,2}$\quad
Runjian Chen$^{4}$\quad
Jinkun Cao$^{5}$\\
Jiangmiao Pang$^{2\textrm{\Letter}}$\quad
Dahua Lin$^{1,2}$\\
$^1$The Chinese University of Hong Kong\quad
$^2$Shanghai AI Laboratory\quad
$^3$S-Lab, NTU\\
$^4$The University of Hong Kong\quad
$^5$Carnegie Mellon University\\
\tt\small \{runsenxu,wt019,dhlin\}@ie.cuhk.edu.hk, wenwei001@ntu.edu.sg, rjchen@connect.hku.hk,\\
\tt\small jinkunc@andrew.cmu.edu, pangjiangmiao@gmail.com}
\maketitle

\blfootnote{\textrm{\Letter} Corresponding author.}

\begin{abstract}

This paper introduces the Masked Voxel Jigsaw and Reconstruction (MV-JAR) method for LiDAR-based self-supervised pre-training and a carefully designed data-efficient 3D object detection benchmark on the Waymo dataset. Inspired by the scene-voxel-point hierarchy in downstream 3D object detectors, we design masking and reconstruction strategies accounting for voxel distributions in the scene and local point distributions within the voxel. We employ a Reversed-Furthest-Voxel-Sampling strategy to address the uneven distribution of LiDAR points and propose MV-JAR, which combines two techniques for modeling the aforementioned distributions, resulting in superior performance. Our experiments reveal limitations in previous data-efficient experiments, which uniformly sample fine-tuning splits with varying data proportions from each LiDAR sequence, leading to similar data diversity across splits. To address this, we propose a new benchmark that samples scene sequences for diverse fine-tuning splits, ensuring adequate model convergence and providing a more accurate evaluation of pre-training methods. Experiments on our Waymo benchmark and the KITTI dataset demonstrate that MV-JAR consistently and significantly improves 3D detection performance across various data scales, achieving up to a 6.3\% increase in mAPH compared to training from scratch. Codes and the benchmark will be available at \href{https://github.com/SmartBot-PJLab/MV-JAR}{https://github.com/SmartBot-PJLab/MV-JAR}.

\end{abstract}
\vspace{-3pt}
\section{Introduction}
\label{sec:intro}
\begin{figure}[t]
\centering
    \includegraphics[width=\linewidth]{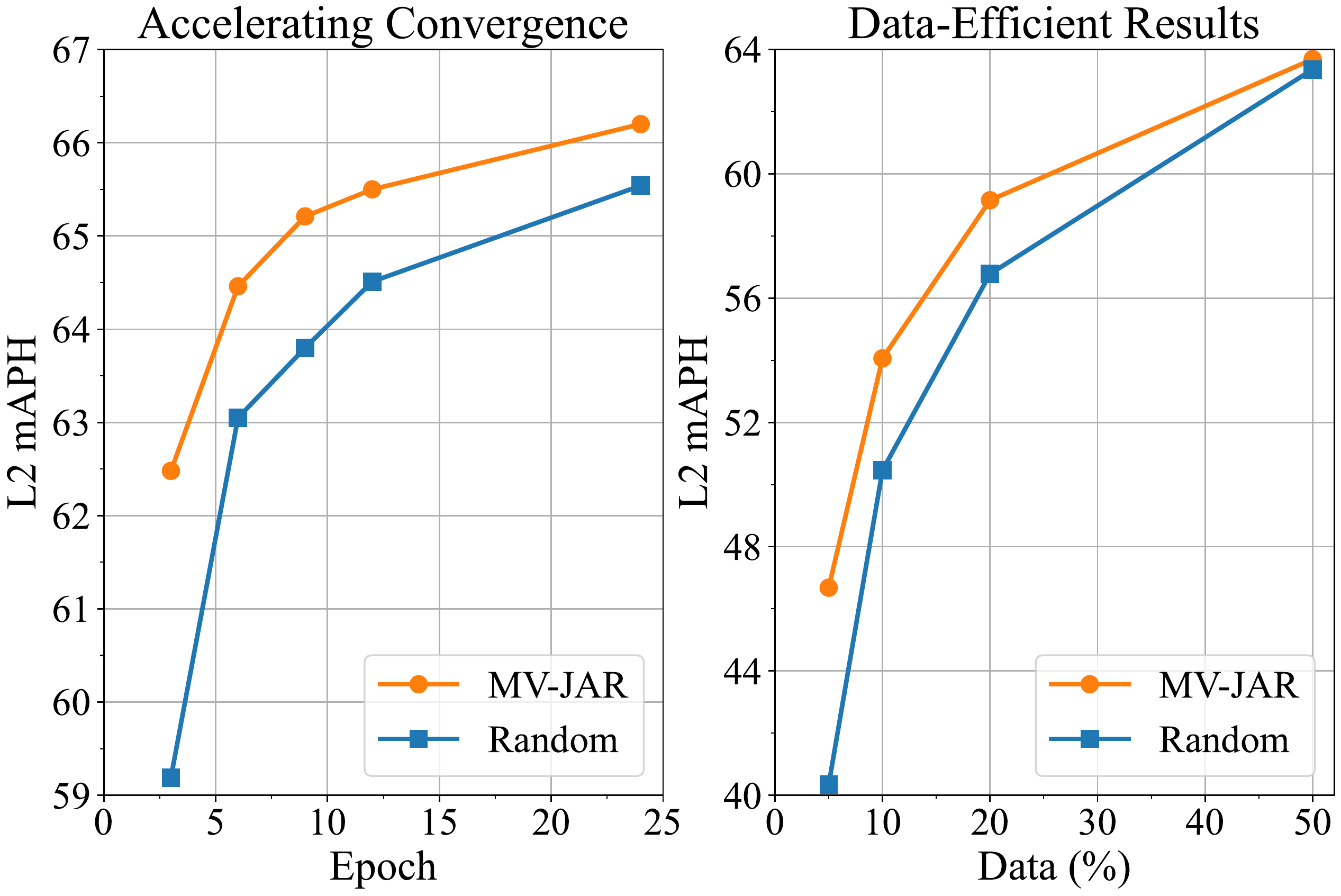}
\caption{3D object detection results on the Waymo dataset. Our MV-JAR pre-training accelerates model convergence and greatly improves the performance with limited fine-tuning data.}
\label{fig:data-efficient_speed_up_convergence}
\vspace{-10pt}
\end{figure}

Self-supervised pre-training has gained considerable attention, owing to its exceptional performance in visual representation learning. Recent advancements in contrastive learning~\cite{moco,mocov2,simclr,swav,zhang2022densesiam} and masked autoencoders~\cite{beit, MAE, SimMiM, ibot,cae} for images have sparked interest among researchers and facilitated progress in modalities such as point clouds.

However, LiDAR point clouds differ from images and dense point clouds obtained by reconstruction as they are naturally sparse, unorganized, and irregularly distributed. Developing effective self-supervised proxy tasks for these unique properties remains an open challenge. Constructing matching pairs for contrastive learning in geometry-dominant scenes is more difficult~\cite{gcc-3d,proposalcontrast}, as points or regions with similar geometry may be assigned as negative samples, leading to ambiguity during training. To address this, our study explores masked voxel modeling paradigms for effective LiDAR-based self-supervised pre-training. 

Downstream LiDAR-based 3D object detectors~\cite{SECOND, centerpoint, pointpillars, SST, reconfig_voxels, ssn} typically quantize the 3D space into voxels and encode point features within them. Unlike pixels, which are represented by RGB values, the 3D space presents a scene-voxel-point hierarchy, introducing new challenges for masked modeling. Inspired by this, we design masking and reconstruction strategies that consider voxel distributions in the scene and local point distributions in the voxel. Our proposed method, Masked Voxel Jigsaw And Reconstruction (MV-JAR), harnesses the strengths of both voxel and point distributions to improve performance.

To account for the uneven distribution of LiDAR points, we first employ a Reversed-Furthest-Voxel-Sampling (R-FVS) strategy that samples voxels to mask based on their sparseness. This approach prevents masking the furthest distributed voxels, thereby avoiding information loss in regions with sparse points. To model voxel distributions, we propose Masked Voxel Jigsaw (MVJ), which masks the voxel coordinates while preserving the local shape of each voxel, enabling scene reconstruction akin to solving a jigsaw puzzle. For modeling local point distributions, we introduce Masked Voxel Reconstruction (MVR), which masks all coordinates of points within the voxel but retains one point as a hint for reconstruction. Combining these two methods enhances masked voxel modeling.

Our experiments indicate that existing data-efficient experiments~\cite{gcc-3d, proposalcontrast} inadequately evaluate the effectiveness of various pre-training methods. The current benchmarks, which uniformly sample frames from each data sequence to create diverse fine-tuning splits, exhibit similar data diversity due to the proximity of neighboring frames in a sequence~\cite{Waymo, KITTI_Det, nuscenes}. Moreover, these experiments train models for the same number of epochs across different fine-tuning splits, potentially leading to incomplete convergence. As a result, the benefits of pre-trained representations become indistinguishable across splits once the object detector is sufficiently trained on the fine-tuning data. To address these shortcomings, we propose sampling scene sequences to form diverse fine-tuning splits and establish a new data-efficient 3D object detection benchmark on the Waymo~\cite{Waymo} dataset, ensuring sufficient model convergence for a more accurate evaluation.

We employ the Transformer-based SST~\cite{SST} as our detector and pre-train its backbone for downstream detection tasks. Comprehensive experiments on the Waymo and KITTI~\cite{KITTI_Det} datasets demonstrate that our pre-training method significantly enhances the model's performance and convergence speed in downstream tasks. Notably, it improves detection performance by 6.3\% mAPH when using only 5\% of scenes for fine-tuning and reduces training time by half when utilizing the entire dataset (Fig.~\ref{fig:data-efficient_speed_up_convergence}). With the representation pre-trained by MV-JAR, the 3D object detectors pre-trained on Waymo also exhibit generalizability when transferred to KITTI.
\section{Related Work}

\noindent\textbf{Self-supervised learning for point clouds.}\quad
Annotating point clouds demands significant effort, necessitating self-supervised pre-training methods. Prior approaches primarily focus on object CAD models~\cite{foldingnet, reconstructsapce, point-mae, point-bert, maskpoint, point-m2ae} and indoor scenes~\cite{PointContrast, scenecontrast, depthcontrast}. Point-BERT~\cite{point-bert} applies BERT-like paradigms for point cloud recognition, while Point-MAE~\cite{point-mae} reconstructs point patches without the tokenizer. To acquire scene-level representation, PointContrast~\cite{PointContrast} introduces a contrastive learning approach that compares points in two static partial views of a reconstructed indoor scene. Its successor, SceneContrast~\cite{scenecontrast}, incorporates spatial information into the contrastive learning framework. However, LiDAR point clouds are irregular and dynamic, with mainstream LiDAR perception models often exhibiting distinct architectures, which obstructs the direct adaptation of object-level and indoor scene pre-training methods. STRL~\cite{strl} employs contrastive learning with two temporally-correlated frames for spatiotemporal representation. GCC-3D~\cite{gcc-3d} presents a framework that combines geometry-aware contrast and pseudo-instance clustering harmonization. ProposalContrast~\cite{proposalcontrast} targets region-level contrastive learning to improve 3D detection. CO3~\cite{co3} leverages infrastructure-vehicle-cooperation point clouds to construct effective contrastive views. In contrast, our work explores masked voxel modeling, diverging from the prevailing contrastive learning paradigm.

\noindent\textbf{Masked autoencoders for self-supervised pre-training.}\quad
Masked language modeling plays a pivotal role in self-supervised pre-training for Transformer-based networks in natural language processing~\cite{bert,gpt-2,gpt-3}. This approach typically masks portions of the input and pre-trains networks to predict the original information. With the successful integration of Transformers into computer vision~\cite{vit}, researchers have increasingly focused on masked image modeling~\cite{beit, MAE, SimMiM, ibot, cae, position_prediction} to mitigate the data-intensive issue of ViT~\cite{vit}. BEiT~\cite{beit} employs a tokenizer to generate discrete tokens for image patches and utilizes a BERT-like framework to pre-train ViT. MAE~\cite{MAE} introduces an asymmetric autoencoder for reconstructing RGB pixels of original images, eliminating the need for an extra tokenizer. SimMiM~\cite{SimMiM} encodes masked raw images with Transformers and employs a lightweight prediction head for recovery. Demonstrating considerable potential in image recognition, masked autoencoders have been applied in video understanding~\cite{Video-MAE,Video-MAEW}, medical image analysis~\cite{mae_for_medical}, and audio processing~\cite{mae_for_audio}. In this work, we investigate the application of this powerful pre-training technique to LiDAR point clouds using voxel representation.

\begin{figure*}[ht]
\centering
\includegraphics[width=\linewidth]{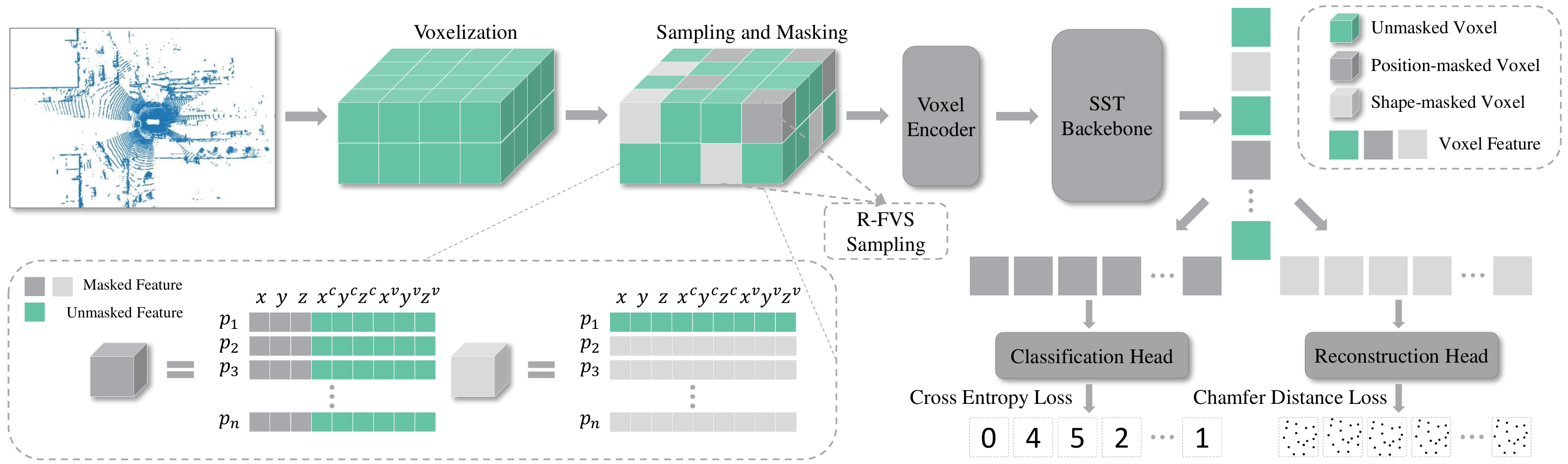}
\caption{Overview of Masked Voxel Jigsaw and Reconstruction (MV-JAR). Our R-FVS sampling method selects a specific ratio of non-empty voxels to mask. For position-masked voxels, we mask the absolute coordinates of all points within the voxel while maintaining the relative coordinates. For shape-masked voxels, we mask both absolute and relative coordinates of all points, preserving only one point. Features are extracted from all non-empty voxels, and masked features are decoded by a dedicated head to recover corresponding targets.}
\label{fig:MV-JAR}
\end{figure*}

\noindent\textbf{Transformer-based 3D object detection.}\quad
The recent success of Vision Transformers~\cite{vit} has inspired extensive research into the application of Transformer-based architectures for 3D object detection~\cite{3DTR, pointformer, voxelformer, SST, SWFormer}. 3DETR~\cite{3DTR} presents an end-to-end object detection framework with modified Transformer blocks for 3D point clouds. Pointformer~\cite{pointformer} employs hierarchical Transformer blocks to extract point features for 3D detection. Voxel Transformer~\cite{voxelformer} develops a Transformer-based 3D backbone to establish long-range relationships between voxels. Recently, SST~\cite{SST} introduced a single-stride sparse Transformer as a replacement for the PointPillars~\cite{pointpillars} backbone, abandoning the multi-resolution strategy to enhance the detection of small objects. Emulating the Swin-Transformer~\cite{swin-transformer}, SST partitions the space into windows and calculates attention only within each window for efficiency. This approach demonstrates impressive detection results, and we adopt it for our experiments.
\section{Methodology}

In this section, we first present an overview of a general masked voxel modeling framework for LiDAR point clouds and our Reversed-Furthest-Voxel-Sampling strategy to tackle the uneven distribution of LiDAR points (\cref{subsec:MVM}).
Masked Voxel Jigsaw (MVJ) (\cref{subsec:MVJ}) and Masked Voxel Reconstruction (MVR) (\cref{subsec:MVR}) are both instantiations of the general framework that learns the voxel distributions and local point distributions, respectively.
Masked Voxel Jigsaw and Reconstruction (MV-JAR) consist of these two tasks to synergically learn the data distributions at two levels (\cref{subsec:joint_pre_training}).

\subsection{Masked Voxel Modeling and Sampling}
\label{subsec:MVM}
As illustrated in \cref{fig:MV-JAR}, a general masked voxel modeling framework converts LiDAR point clouds into voxels and samples a proportion of non-empty voxels using a specific sampling strategy and masking ratio. Various features within the voxel can be masked for a particular learning target in this general framework to instantiate different self-supervised learning tasks. Specifically, task-specific features of the points in the sampled voxels are replaced with a learnable vector.
Subsequently, all non-empty voxels are encoded by a voxel encoder, processed by a backbone network to extract features, and followed by a lightweight decoder that reconstructs task-specific targets from each voxel feature. Each stage is described in the following sections.
In this paper, we employ SST~\cite{SST}, which demonstrates superior performance in 3D detection.
The pre-trained weights of the voxel encoder and the SST backbone are utilized as initialization for the downstream 3D detection task.

\noindent\textbf{Voxelization.}\quad Let $p$ denote a point with coordinates $x, y, z$. Given a point cloud $ P = \{ p_{i} = [x_i, y_i, z_i]^{T} \in \mathbb{R}^{3} \}_{i=1...n} $ containing $n$ points and range $W, H, D$, we partition the space into a 3D grid using voxel sizes $v_W, v_H, v_D$ along the three axes, grouping points in the same voxel together. As a common technique~\cite{SST, pointpillars, voxelnet}, we decorate each point with $x^c, y^c, z^c, x^v, y^v, z^v$, where the superscript $c$ denotes distance to the clustering center of all points in each voxel and the superscript $v$ denotes distance to the voxel center.
Considering the sparse and irregular nature of LiDAR point clouds, we focus only on non-empty voxels. Assuming non-empty voxels contain at most $T$ points and a total of $N$ non-empty voxels, we obtain the voxel-based point cloud representation as $M = \{V_i \}_{i=1...N}$, where $V_i = \{p_j = [x_j, y_j, z_j, x^c_j, y^c_j, z^c_j, x^v_j, y^v_j, z^v_j]^{T} \in \mathbb{R}^{9}\}_{j=1...t_i}$ and $t_i \leq T$. We assume each non-empty voxel contains $T$ points for simplicity. We also maintain each voxel's coordinate in the 3D grid with $ C_{i} \in \mathbb{R}^{3}$.

\noindent\textbf{Masked voxel sampling.}\quad
A key component of masked autoencoders is the mask sampling strategy. Unlike 2D pixels that uniformly distribute on the image plane, voxels are unevenly distributed due to the sparse and irregular nature of LiDAR point clouds. Random sampling may result in information loss in regions with sparse points. Inspired by furthest point sampling~\cite{pointnet_plus} (FPS), which samples an evenly distributed point set while maintaining the key structure of the data, we propose a Reversed-Furthest-Voxel-Sampling (R-FVS) scheme. Given a masking ratio $r$, we apply FPS to select $\lfloor N(1-r) \rfloor$ voxels from all non-empty voxels according to each voxel's coordinate $C_i$. These sampled voxels are \emph{kept} to preserve points in sparse regions and maintain key geometric structures in the data, while the remaining unsampled voxels are masked during training.

\noindent\textbf{Masking voxels.}\quad We directly mask the raw inputs of voxels, providing a unified formulation for the masked voxel modeling framework. For each of the remaining unsampled $ \lceil Nr \rceil $ voxels to be masked, we replace some of the original features of each point in the voxel with a shared learnable vector, denoted as a mask token $ m $. The dimensions of $m$ and the replaced point features vary for different tasks.

\noindent\textbf{Voxel encoding and decoding.}\quad
All the non-empty voxels are then encoded by a voxel encoder (VE) and the SST backbone. As the features of masked voxels are also extracted by the backbone, we only need a lightweight task-specific MLP head to decode and predict their original information~\cite{SimMiM}. The loss is computed only on masked voxels, following previous practices~\cite{MAE, SimMiM}.

\subsection{Masked Voxel Jigsaw (MVJ)}\label{subsec:MVJ}
Object detection in LiDAR point clouds necessitates a representation capable of capturing contextual information, such as the distribution and relationships among voxels. We propose a jigsaw puzzle that obscures position information, compelling the model to learn these relationships.

\noindent\textbf{Masking position information.}\quad
Given a position-masked voxel, we replace the absolute coordinates $x_j, y_j, z_j$ of each point with a mask token $m_v \in \mathbb{R}^{3}$, while preserving the local shape by retaining the decorated coordinates of each point within the voxel (Fig.~\ref{fig:MV-JAR}). The masked voxel is then represented as $V_{\text{masked}} = \{p_j=[m_v, x^c_j, y^c_j, z^c_j, x^v_j, y^v_j, z^v_j]^T \in \mathbb{R}^{9} \}_{j=1...T}$. All voxels are subsequently input to the voxel encoder and the SST backbone. Notably, positional embeddings are not added to the voxels before being fed to the backbone, as the model is tasked with predicting position information.

\noindent\textbf{Prediction target and loss function.}\quad
Instead of directly predicting the absolute position of the masked voxels, we formulate a classification task to facilitate optimization. Owing to the extensive range of LiDAR point clouds, the SST backbone partitions the 3D voxelized grid into windows, akin to the Swin-transformer~\cite{swin-transformer}, to enable efficient attention calculation within windows. We employ the same window partitioning method, requiring the model to predict only the relative index of the masked voxel within its corresponding window. Assuming a 3D window contains $N_x, N_y, N_z$ voxels along the $x, y, z$ axes and the voxel coordinates of a masked voxel are $X, Y, Z$, the relative coordinates $I_x, I_y, I_z$ of the masked voxel with respect to its window are calculated as $I_x = X \mod N_x, I_y = Y \mod N_y, I_z = Z \mod N_z$. The relative index is then given by $I = I_x + I_yN_x + I_zN_xN_y$. The prediction head outputs a classification vector $\hat{v} \in \mathbb{R}^{N_xN_yN_z}$ representing the probabilities of the masked voxel occupying each position. The prediction loss for all masked voxel is calculated using the Cross-Entropy loss as follows:
\begin{equation}
L_{MVJ} = \frac{1}{R_p}\sum_{i=1}^{R_p}\text{CrossEntropy}(\hat{v}_i, I_i),
\end{equation}
where $R_p$ denotes the number of position-masked voxels.

\subsection{Masked Voxel Reconstruction (MVR)}\label{subsec:MVR}
To incorporate local shape information within each voxel, we introduce masked voxel reconstruction (MVR). MVR masks the absolute and relative coordinates of all points in shape-masked voxels to learn point distributions, preserving only one point to provide the voxel position.

\noindent\textbf{Masking voxel shapes.}\quad
In a given voxel, both the original and decorated coordinates of points represent the local shape of the voxel (Fig.~\ref{fig:MV-JAR}). We employ a shared learnable token $m_{p} \in \mathbb{R}^{9}$ to replace all point features in every shape-masked voxel, except for one point. This particular point conveys the voxel's positional information without revealing the voxel shape, which would render the reconstruction task trivial. Consequently, the masked voxel becomes $V_{\text{masked}} = \{[x_1, y_1, z_1, x^c_1, y^c_1, z^c_1, x^v_1, y^v_1, z^v_1]\} \cup \{p_j = m_p \}_{j=2...T}$.

\noindent\textbf{Reconstruction target and loss function.}\quad
We reconstruct each masked voxel and use the $L2$ Chamfer Distance loss to measure the discrepancy between the reconstructed and target point distributions. This loss function is insensitive to point density~\cite{balancedCD}, which is crucial since each voxel may contain varying numbers of points. Assuming each reconstructed voxel contains $n$ points, the reconstruction head outputs a vector containing the coordinates of each point in the masked voxel, $\hat{v} \in \mathbb{R}^{3n}$. To pre-train the model more stably, we normalize the target point distribution using each point's distance to the voxel center as ground truth and scale the distance between 0 and 1. Supposing each reconstructed masked voxel is $\hat{V} = \{ \hat{p}_j = [\hat{x}_j, \hat{y}_j, \hat{z}_j] \}_{j=1...n}$ and the ground truth is $V = \{p_j = [x^v_j, y^v_j, z^v_j] \}_{j=1...t_i}$, the total loss across all masked voxels is given by:
\begin{equation}
L_{MVR} = \frac{1}{R_s}\sum_{i=1}^{R_s}{L_{CD}}(\hat{V}_i, V_i ),
\end{equation}
where $R_{s}$ represents the number of shape-masked voxels and $L_{CD}$ is the $L2$ Chamfer Distance loss calculated as:
\begin{equation}
\begin{split}
        L_{CD} & = \frac{1}{\left|\hat{V}_i\right|} \sum_{\hat{p} \in \hat{V}_i} \min _{p \in V_i}\|\hat{p} - p\|_{2}^{2} \\
        & + \frac{1}{\left|V_i\right|} \sum_{p \in V_i} \min _{\hat{p} \in \hat{V}_i}\|p-\hat{p}\|_{2}^{2}.
\end{split}
\end{equation}

\subsection{Joint Pre-Training}
\label{subsec:joint_pre_training}
We jointly pre-train the model using both MVJ and MVR tasks to enable the model to learn both point and voxel distributions. We use R-FVS to sample $R_s$ voxels for MVR and $R_p$ voxels for MVJ. As illustrated in \cref{fig:MV-JAR}, the voxel encoder and the backbone extract features from both types of masked voxels and unmasked voxels. Masked features are decoded to recover their corresponding targets, and Cross-Entropy loss and Chamfer Distance loss are utilized for supervision. The total loss for joint pre-training is expressed as follows:
\begin{equation}
L = \alpha L_{MVJ} + \beta L_{MVR}
\end{equation}
where $\alpha$ and $\beta$ are balancing coefficients.

\begin{table*}
\centering
\caption{Data-efficient 3D object detection results of SST on the Waymo validation set. SST is pre-trained using various self-supervised methods on Waymo training split, with weights applied for the downstream detection task. ``Random initialization" denotes training from scratch. The model performances are shown using different amounts and diversities of fine-tuning data.}
\label{tab:data-efficient_benchmark_waymo}
\scalebox{0.85}{\tablestyle{8pt}{1.0}
\begin{tabular}{@{}c|c|cc|cccccc@{}}
\toprule
\multirow{2}{*}{Data amount} & \multirow{2}{*}{Initialization} & \multicolumn{2}{c|}{Overall} & \multicolumn{2}{c}{Car} & \multicolumn{2}{c}{Pedestrian} & \multicolumn{2}{c}{Cyclist} \\ \cmidrule(l){3-10} 
 &  & L2 mAP & L2 mAPH & L2 mAP & L2 mAPH & L2 mAP & L2 mAPH & L2 mAP & L2 mAPH \\ \midrule
\multirow{4}{*}{5\%} & Random & 44.41 & 40.34 & 51.01 & 50.49 & 52.74 & 42.26 & 29.49 & 28.27 \\
 & PointContrast\cite{PointContrast} & 45.32 & 41.30 & 52.12 & 51.61 & 53.68 & 43.22 & 30.16 & 29.09 \\
 & ProposalContrast\cite{proposalcontrast} & 46.62 & 42.58 & 52.67 & 52.19 & 54.31 & 43.82 & 32.87 & 31.72 \\
 & \textbf{MV-JAR (Ours}) & \textbf{50.52}\improve{6.11} & \textbf{46.68}\improve{6.34} & 56.47 & 56.01 & 57.65 & 47.69 & 37.44 & 36.33 \\ \midrule
\multirow{4}{*}{10\%} & Random & 54.31 & 50.46 & 54.84 & 54.37 & 60.55 & 50.71 & 47.55 & 46.29 \\
 & PointContrast\cite{PointContrast} & 53.69 & 49.94 & 54.76 & 54.30 & 59.75 & 50.12 & 46.57 & 45.39 \\
 & ProposalContrast\cite{proposalcontrast} & 53.89 & 50.13 & 55.18 & 54.71 & 60.01 & 50.39 & 46.48 & 45.28 \\
 & \textbf{MV-JAR (Ours)} & \textbf{57.44}\improve{3.13} & \textbf{54.06}\improve{3.60} & 58.43 & 58.00 & 63.28 & 54.66 & 50.63 & 49.52 \\ \midrule
\multirow{4}{*}{20\%} & Random & 60.16 & 56.78 & 58.79 & 58.35 & 65.63 & 57.04 & 56.07 & 54.94 \\
 & PointContrast\cite{PointContrast} & 59.35 & 55.78 & 58.64 & 58.18 & 64.39 & 55.43 & 55.02 & 53.73 \\
 & ProposalContrast\cite{proposalcontrast} & 59.52 & 55.91 & 58.69 & 58.22 & 64.53 & 55.45 & 55.36 & 54.07 \\
 & \textbf{MV-JAR (Ours)} & \textbf{62.28}\improve{2.11} & \textbf{59.15}\improve{2.37} & 61.88 & 61.45 & 66.98 & 59.02 & 57.98 & 57.00 \\ \midrule
\multirow{4}{*}{50\%} & Random & 66.43 & 63.36 & 63.81 & 63.38 & 70.78 & 63.05 & 64.71 & 63.66 \\
 & PointContrast\cite{PointContrast} & 65.51 & 62.21 & 62.66 & 62.23 & 69.82 & 61.53 & 64.04 & 62.86 \\
 & ProposalContrast\cite{proposalcontrast} & 65.76 & 62.49 & 62.93 & 62.50 & 70.09 & 61.86 & 64.26 & 63.11 \\
 & \textbf{MV-JAR (Ours)} & \textbf{66.70}\improve{0.27} & \textbf{63.69}\improve{0.33} & 64.30 & 63.89 & 71.14 & 63.57 & 64.65 & 63.63 \\ \midrule
\multirow{4}{*}{100\%} & Random & 68.50 & 65.54 & 64.96 & 64.56 & 72.38 & 64.89 & 68.17 & 67.17 \\
 & PointContrast\cite{PointContrast} & 68.06 & 64.84 & 64.24 & 63.82 & 71.92 & 63.81 & 68.03 & 66.89 \\
 & ProposalContrast\cite{proposalcontrast} & 68.17 & 65.01 & 64.42 & 64.00 & 71.94 & 63.94 & 68.16 & 67.10 \\
 & \textbf{MV-JAR (Ours)} & \textbf{69.16}\improve{0.66} & \textbf{66.20}\improve{0.66} & 65.52 & 65.12 & 72.77 & 65.28 & 69.19 & 68.20 \\ \bottomrule
\end{tabular}
}
\end{table*}
\section{Data-Efficient Benchmark on Waymo}
\label{sec:benchmark}

To evaluate the pre-trained representation, we transfer the weights of the voxel encoder and the backbone for object detection. Previous benchmarks fail to effectively reveal the effectiveness of pre-training strategies due to data diversity issues. We propose a new data-efficient benchmark to address these limitations.

\begin{figure}
\centering
        \begin{subfigure}{0.49\linewidth}
            \includegraphics[width=\linewidth]{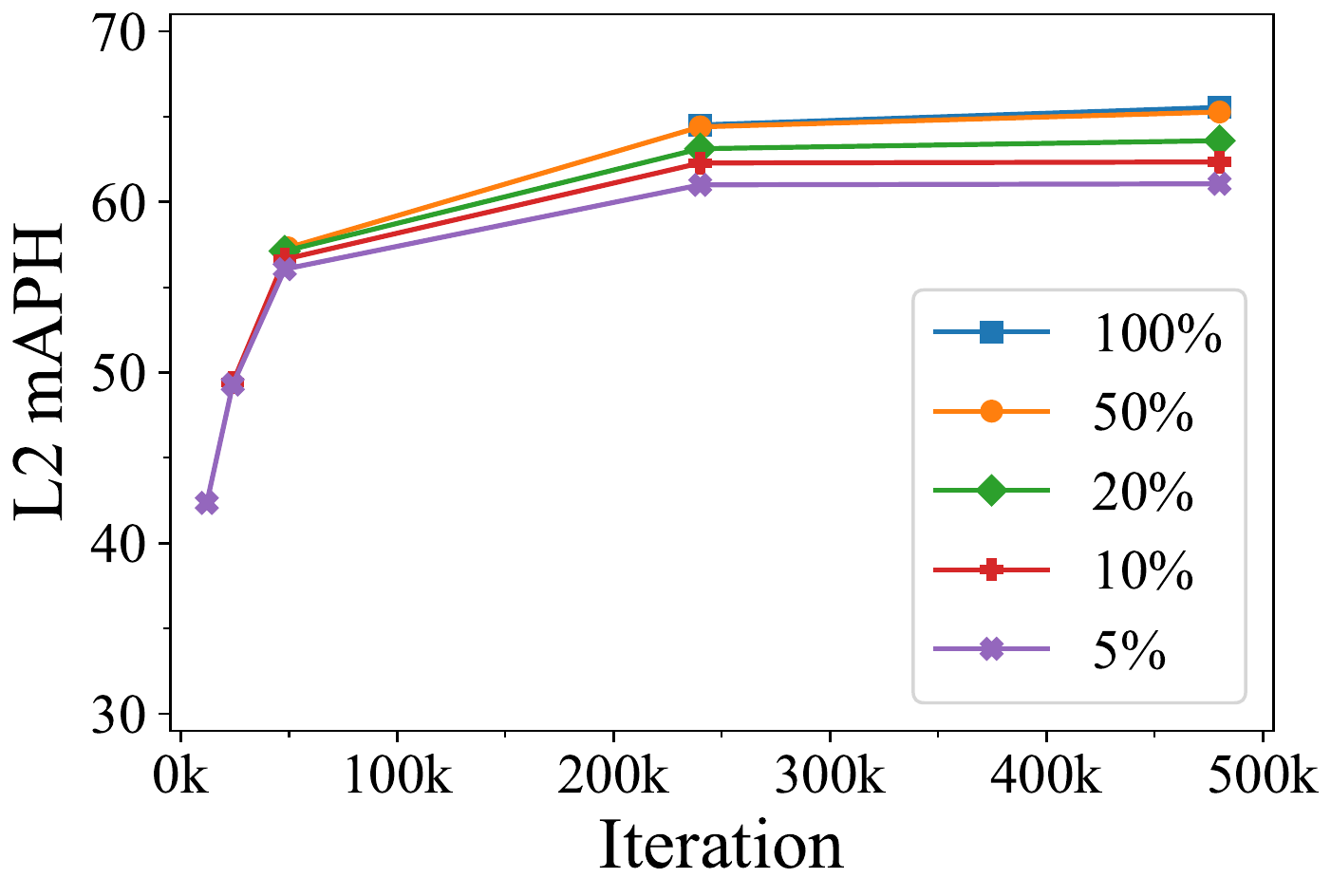}
        \caption{Existing}
        \label{fig:existing_benchmark}
        \end{subfigure}
        \hfill
        \begin{subfigure}{0.49\linewidth}
            \includegraphics[width=\linewidth]{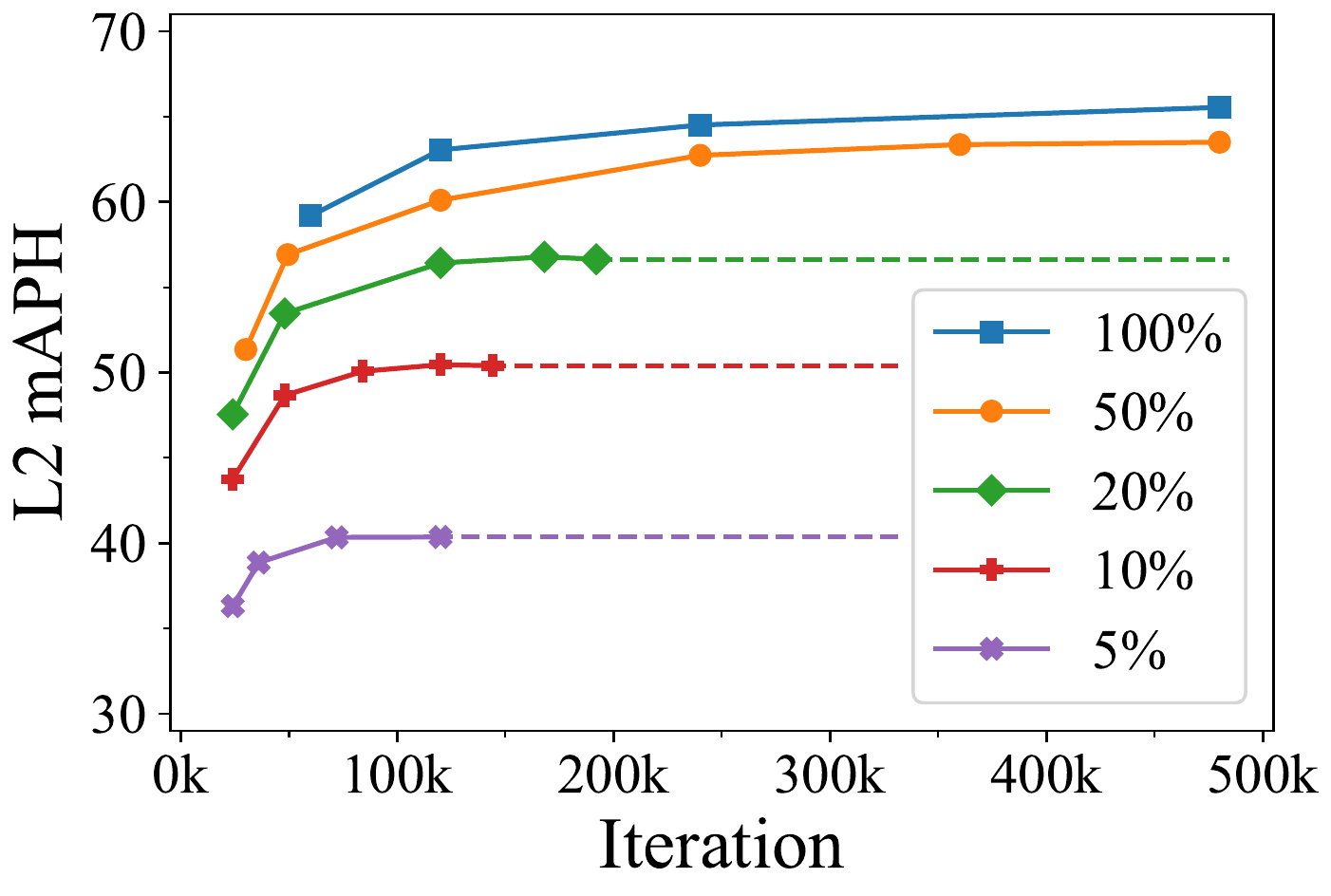}
        \caption{Ours}
        \label{fig:our_benchmark}
        \end{subfigure}

\caption{Comparison of data-efficient benchmarks. Previous uniformly sampled splits exhibit similar diversity, as evidenced by similar performance with the same training iterations. Our benchmark features varying data amounts and diversities, ensuring model convergence. Dashed lines show baseline performances.}
\label{partial-ft}
\end{figure}

\noindent\textbf{Observation: Incomplete model convergence.}\quad
Prior works~\cite{gcc-3d, proposalcontrast} fine-tune the model on different splits with the same \emph{epochs}, causing reduced iteration numbers when dataset sizes decrease. As illustrated in \cref{fig:existing_benchmark}, by training for more epochs (but the same \emph{iterations}), the performance of SST significantly improves, suggesting that the models may not fully converge on smaller splits when training epochs remain constant. This entangled factor prevents accurate evaluation of pre-training strategies.

\noindent\textbf{Issue: Fine-Tuning splits with similar data diversities.}\quad
With the same training iterations, models trained on different splits display comparable performance, despite significant differences in data amounts (\cref{fig:existing_benchmark}). We find that scene diversity is a crucial factor. Previous works \cite{gcc-3d, proposalcontrast} uniformly sample the entire Waymo training set to create different fine-tuning splits, for instance, sampling one frame from every two consecutive frames for a 50\% split. However, neighboring frames in self-driving datasets are similar due to the short time difference between consecutive frames (e.g., only 0.1$s$ in Waymo~\cite{Waymo}). This results in similar scene diversity across splits, leading to comparable detection performance using the same training iterations. 

We argue that such an experimental setting does not accurately represent a typical application scenario of self-supervised pre-training, where annotated fine-tuning data is less abundant and diverse compared to pre-training data. Furthermore, it fails to evaluate how pre-training benefits downstream tasks when varying amounts and diversities of labeled data are available. We provide an additional example in the supplementary material to further illustrate these issues with previous benchmarks. Our proposed data-efficient benchmark addresses these issues, offering a more precise evaluation of pre-training strategies.

\noindent\textbf{Solution: Sequence-Based data sampling.}\quad
We sample fine-tuning data using different scene sequences rather than uniformly sampling varying numbers of frames from identical sequences. The Waymo training split comprises 798 distinct scene sequences, each lasting 20$s$ with approximately 200 frames\cite{Waymo}. We randomly sample {5\%, 10\%, 20\%, 50\%} of the scene sequences, respectively, using all frames within the sampled sequences for fine-tuning. Consequently, data diversities and amounts vary across each split. Each larger split also encompasses the smaller ones, enabling measurement of performance changes with additional fine-tuning data. Furthermore, we randomly sample three 5\% and 10\% splits and report average results to reduce variance on these smaller sets (see our supplementary material).

For different fine-tuning splits, we train the model until performance saturates (i.e., overfitting) and determine the number of epochs used for experiments accordingly. For example, for 5\% data, we try \{48, 60, 66, 72, 78, 84\} epochs and ultimately select 72. The baseline performances across various training iterations are shown in \cref{fig:our_benchmark}. With our new benchmark, baseline performance and convergence speed vary as anticipated for different splits.

\section{Experiments}

\subsection{Experimental Settings}

\noindent\textbf{Datasets.}\quad The Waymo dataset~\cite{Waymo} consists of 1150 self-driving scenes with 798 scenes as the training split, 202 scenes as the validation split, and 150 scenes as the testing split. We pre-train on the training split, which contains 158,240 annotated frames. For the downstream data-efficient 3D object detection task, we fine-tune the model using our sampled splits (as described in \cref{sec:benchmark}) and evaluate performance on the standard Waymo validation split with 3 classes (cars, pedestrians, and cyclists) and two difficulty levels (L1 and L2). We adopt L2 mean average precision (L2 mAP) and L2 mean average precision with heading (L2 mAPH) as main evaluation metrics. We also evaluate the model on the KITTI dataset\cite{KITTI_Det} for the detection task. It contains 7481 training samples and 7518 test samples.

\noindent\textbf{Implementation details.}\quad We employ the official implementation and training settings of SST\cite{SST} for pre-training and fine-tuning unless specified otherwise. The Waymo dataset's point cloud range is $W \times H \times D=149.76m \times 149.76m \times 6m$, and the voxel size is $v_W \times v_H \times v_D = 0.32m \times 0.32m \times 6m$. For MVR, we predict 15 points per voxel with a masking ratio of 0.05. For MVJ, the window size is $N_x \times N_y \times N_z = 12 \times 12 \times 1$ and the masking ratio is 0.1. We set the loss weight $\alpha=\beta=1$, pre-train for 6 epochs with an initial learning rate of $5e-6$. When fine-tuning, we initialize the voxel encoder and SST backbone with pre-trained weights, and other training hyper-parameters remain unchanged. We set KITTI's point cloud range to $69.12m \times 79.36m \times 4m$, voxel size to $0.32m \times 0.32m \times 4m$, and train for 160 epochs. We use the AdamW optimizer with a batch size of 8 and a cyclic learning rate scheduler with cosine annealing for all training.

\subsection{Main Results on Waymo}
\noindent\textbf{Baselines.}\quad We fine-tune the model using data splits described in \cref{sec:benchmark}. For \{5\%, 10\%, 20\%, 50\%, 100\%\} data, the model is trained for \{72, 60, 42, 36, 24\} epochs. We use the randomly initialized model's performances as our baselines. We also include experiments with models using convolutional backbones in the supplementary material.

\noindent\textbf{Results with different initialization.}\quad \cref{tab:data-efficient_benchmark_waymo} presents the SST model's performances with various initializations on our proposed data-efficient benchmark. Our pre-training method consistently improves SST baselines across all fine-tuning data amounts, especially when data is scarce. Our method achieves up to 15.7\% relative improvement, from 40.34\% L2 mAPH to 46.68\% L2 mAPH, when using only 5\% fine-tuning data. The performance gain exists across object categories, indicating effective general representation learning. As more fine-tuning data is introduced, the performance gain persists, and when fine-tuning with the entire Waymo training set, our pre-training enhances detection performance from 65.54\% mAPH to 66.20\% mAPH.

We note that with 50\% and 100\% fine-tuning data, the improvements from our pre-training are not as significant as with less data. We observe that the baseline result on 50\% data (63.36\% mAPH) is comparable to that on 100\% data (65.54\% mAPH), suggesting that training data is no longer the bottleneck. The original SST, with only 2.1M parameters, is a lightweight network. We scale up SST to 8.3M parameters by using larger hidden layer channels and observe larger benefits from pre-training on the 50\% fine-tuning split, from 63.13\% L2 mAPH to 64.24\% L2 mAPH. Therefore, we argue that the SST model's capacity, rather than the fine-tuning data amount, becomes the bottleneck for detection performance.

\noindent\textbf{Comparisons with contrastive learning methods.}\quad We utilize the official implementation of the recently proposed ProposalContrast~\cite{proposalcontrast} and reimplement PointContrast~\cite{PointContrast} for comparison. As demonstrated in \cref{tab:data-efficient_benchmark_waymo}, both contrastive learning methods offer benefits only with 5\% fine-tuning data, and their improvements diminish with higher baseline performances. This may be due to the difficulty in smoothly learning representations. A key challenge in LiDAR-based contrastive learning is constructing matched pairs. The point pairs used by PointContrast face the issue of spatially adjacent points with similar features being assigned as negative samples. We observe difficulty in convergence when sampling excessive pairs. ProposalContrast alleviates the issue by contrasting regions but does not entirely resolve it. Alternatively, our method offers a new avenue for developing self-supervised pre-training through masked voxel modeling on LiDAR point clouds.

\subsection{Pre-training Accelerates Convergence}
To investigate how MV-JAR pre-training accelerates convergence, we fine-tune the SST model on the entire Waymo training split. We train the SST model for different epochs and report the performance of both fine-tuning and training from scratch. As illustrated in \cref{fig:data-efficient_speed_up_convergence}, MV-JAR pre-training significantly enhances the convergence speed of SST. By fine-tuning for only 3 epochs, the SST model achieves 62.48\% L2 mAPH. Remarkably, fine-tuning for just 12 epochs results in a performance comparable to training from scratch for 24 epochs, effectively halving the convergence time. As the number of fine-tuning epochs increases, the performance gain provided by MV-JAR does not wane. Pre-training can still improve the model's converged performance by 0.74\% L2 mAPH, highlighting the superiority of the pre-trained representation.

\subsection{Transferring Results on KITTI} 
\begin{table}
\begin{minipage}{\linewidth}
    \centering
    \caption{Transferring results on the KITTI validation split. We pre-train SST with the Waymo training split and fine-tune with the KITTI training split.}
    \label{tab:transfer_kitti}
    \scalebox{0.85}{\tablestyle{8pt}{1.0}
    \begin{tabular}{@{}cccc@{}}
    \toprule
    \multirow{2}{*}{Initialization} & \multicolumn{3}{c}{Overall} \\ \cmidrule(l){2-4} 
     & Easy & Mod. & Hard \\ \midrule
    Random & 74.71 & 63.43 & 60.00 \\
    PointContrast\cite{PointContrast}  & 73.35 & 62.53 & 59.01 \\
    ProposalContrast\cite{proposalcontrast}  & 73.63 & 63.34 & 59.40 \\
    \textbf{MV-JAR (Ours)} & \textbf{75.22} & \textbf{63.80} & \textbf{60.35} \\ \bottomrule
    \end{tabular}}
\end{minipage}
\end{table}

To evaluate the transferability of the learned representation, we fine-tune various pre-trained SST models on the KITTI dataset. The original SST paper\cite{SST} did not conduct experiments on the KITTI dataset; thus, we follow the training schedule and setup of SST and PointPillars\cite{pointpillars} within the MMDetection3D framework\cite{mmdet3d2020}.
As demonstrated in \cref{tab:transfer_kitti}, the performance gain from our MV-JAR pre-training persists when the representation is transferred to a different domain, indicating that the model learns a generic representation through pre-training. While the KITTI training samples account for approximately 5\% of the entire Waymo training split, the relative improvement is much smaller than transferring to the 5\% Waymo split. We hypothesize that this may be attributed to the domain gap between the Waymo and KITTI datasets.

\subsection{Ablation Studies}   
\begin{figure*}[t]
  \centering
    \includegraphics[width=\linewidth]{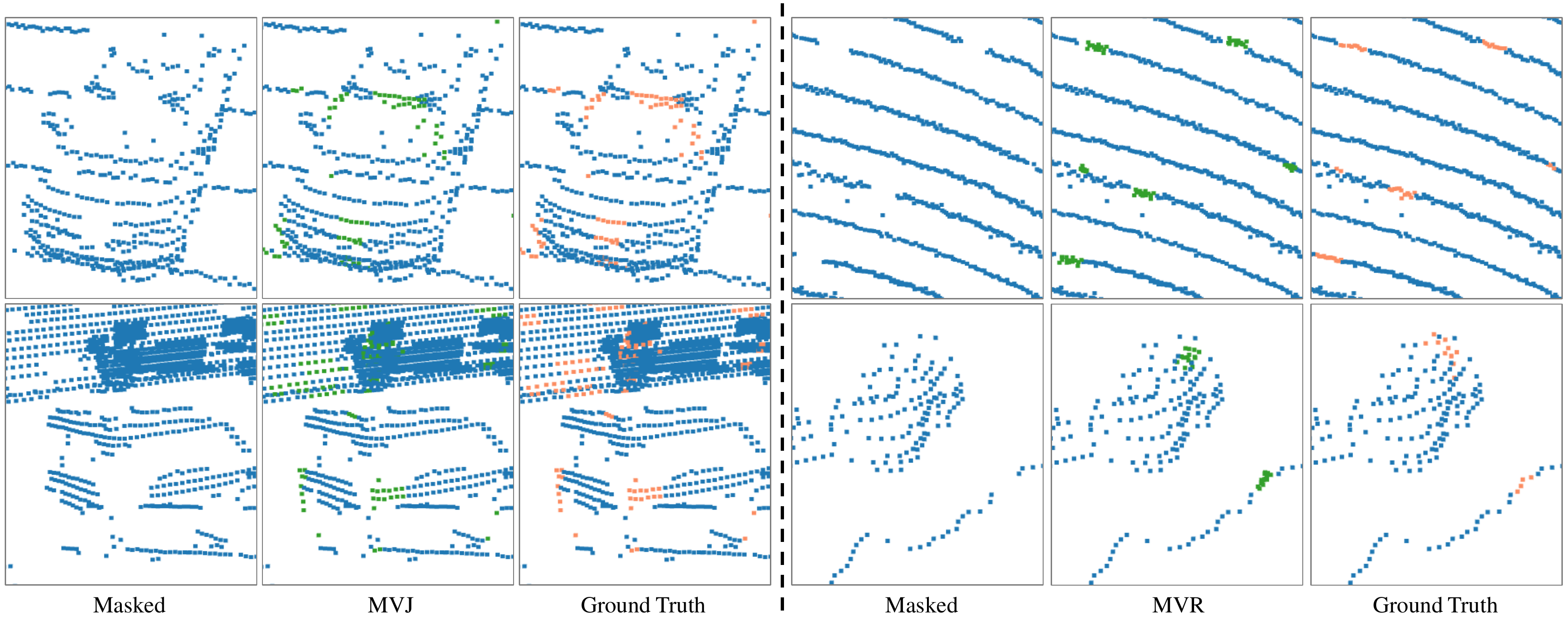}
  \caption{Visualization of the reconstruction results of MV-JAR on the Waymo validation set, compared with ground truths.}
  \label{fig:visualization}
\end{figure*}

\begin{table*}
\small
\begin{minipage}{.3\linewidth}
    \centering
\caption{Ablation study on mask sampling strategies.}
\label{tab:sampling_strategy}
    \scalebox{0.85}{\tablestyle{8pt}{1.0}
    \begin{tabular}{@{}ccc@{}}
    \toprule
    \multirow{2}{*}{Mask sampling} & \multicolumn{2}{c}{Overall} \\ \cmidrule(l){2-3} 
     & L2 mAP & L2 mAPH \\ \midrule
    FVS & 49.73 & 45.88 \\
    Random & 49.86 & 46.06 \\
    \textbf{R-FVS} & \textbf{50.52} & \textbf{46.68} \\ \bottomrule
    \end{tabular}}
\end{minipage}
\begin{minipage}{.4\linewidth}
\centering
\caption{Ablation study on masking ratios.}
\label{tab:mask_ratio}
    \scalebox{0.85}{\tablestyle{8pt}{1.0}
    \begin{tabular}{@{}cccc@{}}
    \toprule
    \multicolumn{2}{c}{Masking ratio} & \multicolumn{2}{c}{Overall} \\ \midrule
    MVJ & MVR & L2 mAP & L2 mAPH \\ \midrule
    0.3 & 0.1 & 49.74 & 46.05 \\
    0.05 & 0.1 & 49.67 & 45.92 \\
    0.1 & 0.1 & 49.85 & 45.97 \\
    \textbf{0.1} & \textbf{0.05} & \textbf{50.52} & \textbf{46.68} \\
    0.1 & 0.3 & 50.25 & 46.35 \\ \bottomrule
    \end{tabular}}
\end{minipage}
\begin{minipage}{.3\linewidth}
\centering
\caption{Ablation study on pre-training effects of MVR and MVJ.}
\label{tab:Pre-training_Effects}
    \scalebox{0.85}{\tablestyle{8pt}{1.0}
    \begin{tabular}{@{}ccc@{}}
    \toprule
    \multirow{2}{*}{Task} & \multicolumn{2}{c}{Overall} \\ \cmidrule(l){2-3} 
     & L2 mAP & L2 mAPH \\ \midrule
    MVR & 46.57 & 42.73 \\
    MVJ & 49.96 & 46.10 \\
    \textbf{MV-JAR} & \textbf{50.52} & \textbf{46.68} \\ \bottomrule
    \end{tabular}}
\end{minipage}
\end{table*}

In this section, we conduct ablation studies to investigate the effectiveness of our proposed Reversed-Furthest-Voxel-Sampling (R-FVS) strategy. Additionally, we explore the optimal masking ratio and the impact of MVR and MVJ pre-training. All pre-training experiments utilize the entire Waymo training split, and we fine-tune the model on one of our 5\% data splits.

\noindent\textbf{Mask sampling strategy.}\quad 
Our R-FVS mask sampling strategy masks voxels that are not sampled by FPS, aiming to avoid masking voxels located in sparse regions. We also examine random sampling and an opposing sampling strategy called FVS, which masks voxels sampled by FPS. With FVS sampling, regions with sparse points are more likely to be masked, leading to greater information loss compared to random sampling and R-FVS sampling. \cref{tab:sampling_strategy} presents the fine-tuning results of pre-training with the three sampling strategies. R-FVS sampling performs the best, while FVS performs the worst. These findings confirm that minimizing information loss benefits pre-training and validate the effectiveness of our R-FVS strategy.

\noindent\textbf{Masking ratio.}\quad
\cref{tab:mask_ratio} illustrates the impact of different masking ratio combinations. Specifically, MVR with a 0.1 masking ratio and MVJ with a 0.05 masking ratio yield the best results. With this combination, the overall masking ratio is 0.15, which is significantly lower than the masking ratios in the image\cite{MAE, SimMiM} or video domain\cite{Video-MAE, Video-MAEW}. A high masking ratio in masked autoencoders typically indicates information redundancy\cite{MAE, Video-MAE, Video-MAEW, SimMiM}. We postulate that two factors contribute to our low masking ratio: 1) LiDAR point clouds represent a vast space where each meaningful object occupies only a small fraction. The information on each object is not highly redundant, especially when the objects are distant from the sensor. This contrasts with images in datasets like ImageNet, where objects often encompass a significant portion of the image. 2) The SST model calculates attention only within a partitioned window, which restricts the information used. We leave this question for future exploration.

\noindent\textbf{Pre-training effects of MVR and MVJ.}\quad
We investigate the individual pre-training effects of MVR and MVJ by optimizing the masking ratio for each task and reporting the fine-tuning performances in \cref{tab:Pre-training_Effects}. Both MVR and MVJ pre-training enhance performance compared to random initialization, with MVJ outperforming MVR. This finding suggests that capturing voxel distributions plays a more significant role in representation learning, which is consistent with the fact that LiDAR detectors process points by downsampling them into voxels. Our integrated MV-JAR method demonstrates further improvement over MVR and MVJ, validating the effectiveness of jointly capturing point and voxel distributions. Additionally, we offer an analysis of improvements across various distances in our supplementary material.

\subsection{Visualization}
\Cref{fig:visualization} displays the MV-JAR reconstruction results on the Waymo validation set, including MVJ and MVR outcomes. In our pre-training, MVJ achieves approximately 89\% classification accuracy of masked voxels on the validation set, accurately reconstructing masked voxels in the correct positions most of the time. MVR can capture the point distributions within voxels, evidenced by reconstructed ground points aligning with ground circles. However, MVR struggles to capture detailed distributions, likely due to the discrete sampling of LiDAR points from continuous surfaces, resulting in considerable variability in the data and increased difficulty in capturing finer details.
\section{Conclusions}

In this paper, we introduce the Masked Voxel Jigsaw and Reconstruction (MV-JAR) pre-training method for LiDAR detectors. MV-JAR captures both point and voxel distributions of LiDAR point clouds, enabling models to learn effective and generic representations. We also develop a Reversed-Furthest-Voxel-Sampling strategy to address the uneven distribution of LiDAR points. Comprehensive experiments on the Waymo and KITTI datasets show that our method consistently and significantly enhances the detector's performance across different data scale regimes. MV-JAR offers a promising alternative for LiDAR-based self-supervised pre-training through Masked Voxel Modeling. Additionally, we establish a new data-efficient benchmark on the Waymo dataset, incorporating fine-tuning splits with diverse data variations. This benchmark effectively assesses the impact of pre-training on downstream tasks with varying amounts and diversities of labeled data.

\noindent\textbf{Acknowledgements.}\quad This project is funded in part by the Shanghai AI Laboratory, CUHK Interdisciplinary AI Research Institute, and the Centre for Perceptual and Interactive Intelligence (CPII) Ltd. under the Innovation and Technology Commission (ITC)'s InnoHK.

\begin{figure*}[!t]
\centering
\begin{minipage}[c]{\textwidth}
\centering
\Large\textbf{Supplementary Material}
\end{minipage}
\end{figure*}

\setcounter{section}{0}

\section{Necessity of Our New Benchmark}

To demonstrate the importance of full convergence and our proposed benchmark, we follow previous works~\cite{gcc-3d,proposalcontrast} and fine-tune SST~\cite{SST} on uniformly sampled 5\% data. We conduct fine-tuning for 6 epochs and 84 epochs, respectively. The results in \cref{tab:necessity_of_our_benchmark} reveal that ProposalContrast significantly improves the baseline when the model is trained with few iterations. However, it slightly diminishes performance when the model is fully converged.

The disappearance of pre-training benefits underscores the necessity of adequate fine-tuning. Therefore, fine-tuning the model on different uniformly sampled splits using the same number of \emph{epochs}, which may result in incomplete convergence on smaller splits, cannot precisely assess pre-training effects. In our main paper, we present evidence that uniformly sampled splits are actually similar when the model reaches full convergence. Our proposed benchmark, which samples data by scene sequences to create diverse splits, can effectively and comprehensively reveal the pure improvements of pre-training.

\begin{table}[h]
\begin{minipage}{\linewidth}
    \centering
    \caption{Fine-tuning on uniformly sampled 5\% data.}
    \label{tab:necessity_of_our_benchmark}
    \scalebox{0.8}{\tablestyle{8pt}{1.0}
        \begin{tabular}{@{}ccccc@{}}
        \toprule        
        \multirow{2}{*}{Initialization} & \multicolumn{2}{c}{6 Epochs} & \multicolumn{2}{c}{84 Epochs} \\ \cmidrule(l){2-5} 
         & L2 mAP & L2 mAPH & L2 mAP & L2 mAPH \\ \midrule
        Random & 40.90 & 34.12 & 63.00 & 58.86 \\
        ProposalContrast\cite{proposalcontrast} & 47.56 & 41.30 & 62.57 & 58.75 \\
        \textbf{MV-JAR} & \textbf{50.67} & \textbf{45.02} & \textbf{65.14} & \textbf{61.74} \\ \bottomrule
        \end{tabular}
        }
\end{minipage}
\end{table}
\section{More Results on Waymo Subsets}

To reduce performance variance on our 5\% and 10\% splits of the Waymo~\cite{Waymo} dataset, we randomly sample each split three times to form three subsets. In the main paper, we report the detection results of SST fine-tuned with Subset 0. In this supplementary material, we present the detection results of SST fine-tuned with the other two subsets in \cref{tab:subset1} and \cref{tab:subset2}. Additionally, we report the average results across all three subsets in \cref{tab:average}.

\section{Experiments with Convolution-based Detectors}

\noindent\textbf{Implementation details.}\quad Our MVJ and MVR directly mask the raw inputs of LiDAR point clouds, making them suitable for most 3D detectors that downsample the point clouds into voxels and extract voxel features for perception. However, MVJ aims at predicting the voxel positions, which is necessary for convolutional operations. Directly applying MVJ to convolution-based backbones may result in information leakage and trivial pre-training. This is not an issue with Transformer-based backbones, as the attention mechanism does not require position information to perform, and we do not add positional embeddings during pre-training. On the other hand, MVR predicts voxel shapes while retaining position information, making it compatible with convolution-based detectors without modification.

To overcome the limitation of MVJ when applied to convolution-based detectors, we permutate the masked voxels by randomly placing them in the partitioned window before feeding them to the convolutional backbones. This permutation hides the original position information of the masked voxels, avoiding information leakage and making MVJ pre-training meaningful.

\noindent\textbf{Experimental Results.}\quad To evaluate the performance of our proposed methods on convolution-based detectors, we pre-train PointPillar~\cite{pointpillars} and CenterPoint (Pillar)~\cite{centerpoint} with MVR and MVJ as SST. We fine-tune these models on our 5\% split and report their overall L2 performances in \cref{tab:performances_convolution_backbones}. Our experimental results demonstrate that both MVJ and MVR can work effectively for convolution-based detectors, showcasing the generalization abilities of our methods.

\begin{table}
\begin{minipage}{\linewidth}
    \centering
    \caption{Performances with convolution-based detectors.}
    \label{tab:performances_convolution_backbones}
    \scalebox{0.85}{\tablestyle{8pt}{1.0}
        \begin{tabular}{@{}ccccc@{}}
        \toprule
        \multirow{2}{*}{Initialization} & \multicolumn{2}{c}{PointPillar~\cite{pointpillars}} & \multicolumn{2}{c}{CenterPoint~\cite{centerpoint}} \\ \cmidrule(l){2-5} 
         & L2 mAP & L2 mAPH & L2 mAP & L2 mAPH \\ \midrule
        Random & 41.27 & 35.10 & 36.21 & 32.26 \\
        MVR & 43.82\improve{2.55} & 37.97\improve{2.87} & 38.02\improve{1.81} & 33.73\improve{1.47} \\
        MVJ & 43.01\improve{1.74} & 37.11\improve{2.01} & 38.61\improve{2.40} & 34.48\improve{2.22} \\ \bottomrule
        \end{tabular}
        }
\end{minipage}
\end{table}

\begin{table}[h]
    \centering
    \caption{Overall L2 mAPH across various distances.}
    \label{tab:performance_across_distance}
    \scalebox{0.9}{\tablestyle{8pt}{1.0}
        \begin{tabular}{@{}cccc@{}}
        \toprule
        \multirow{2}{*}{Initialization} & \multicolumn{3}{c}{Overall} \\ \cmidrule(l){2-4} 
         & 0m-30m & 30m-50m & 50m-inf \\ \midrule
        Random & 60.15 & 34.61 & 18.18 \\
        MVR & 62.77 \improve{2.62} & 36.56\improve{1.95} & 19.94\improve{1.76} \\
        MVJ & 65.66\improve{5.51} & 40.61\improve{6.00} & 23.07\improve{4.89} \\
        MV-JAR & 66.95\improve{6.80} & 40.62\improve{6.01} & 22.88\improve{4.70} \\ \bottomrule
        \end{tabular}
    }
\end{table}

\begin{table*}[t]
\centering
\caption{Data-efficient 3D object detection results of SST on the Waymo validation set, fine-tuned with Subset 1.}
\label{tab:subset1}
\scalebox{0.85}{\tablestyle{8pt}{1.0}
\begin{tabular}{@{}c|c|cc|cccccc@{}}
\toprule
\multirow{2}{*}{Data amount} & \multirow{2}{*}{Initialization} & \multicolumn{2}{c|}{Overall} & \multicolumn{2}{c}{Car} & \multicolumn{2}{c}{Pedestrian} & \multicolumn{2}{c}{Cyclist} \\ \cmidrule(l){3-10} 
 &  & L2 mAP & L2 mAPH & L2 mAP & L2 mAPH & L2 mAP & L2 mAPH & L2 mAP & L2 mAPH \\ \midrule
\multirow{4}{*}{5\%} & Random & 47.74 & 43.69 & 50.24 & 49.75 & 51.68 & 41.26 & 41.29 & 40.07 \\
 & PointContrast\cite{PointContrast} & 48.97 & 44.91 & 52.35 & 51.85 & 52.49 & 41.95 & 42.07 & 40.91 \\
 & ProposalContrast\cite{proposalcontrast} & 49.87 & 45.83 & 52.79 & 52.31 & 53.30 & 43.00 & 43.51 & 42.18 \\
 & \textbf{MV-JAR (Ours)} & \textbf{52.73}\improve{4.99} & \textbf{48.99}\improve{5.30} & 56.66 & 56.21 & 57.52 & 47.61 & 44.02 & 43.15 \\ \midrule
\multirow{4}{*}{10\%} & Random & 55.95 & 52.15 & 55.23 & 54.76 & 60.61 & 50.86 & 52.01 & 50.84 \\
 & PointContrast\cite{PointContrast} & 55.22 & 51.31 & 55.62 & 55.15 & 59.25 & 49.17 & 50.81 & 49.60 \\
 & ProposalContrast\cite{proposalcontrast} & 55.59 & 51.67 & 55.57 & 55.12 & 60.02 & 49.98 & 51.18 & 49.90 \\
 & \textbf{MV-JAR (Ours)} & \textbf{58.61}\improve{2.66} & \textbf{55.12}\improve{2.97} & 58.92 & 58.49 & 63.44 & 54.40 & 53.48 & 52.47 \\ \bottomrule
\end{tabular}
}
\end{table*}

\begin{table*}[h]
\centering
\caption{Data-efficient 3D object detection results of SST on the Waymo validation set, fine-tuned with Subset 2.}
\label{tab:subset2}
\scalebox{0.85}{\tablestyle{8pt}{1.0}
\begin{tabular}{@{}c|c|cc|cccccc@{}}
\toprule
\multirow{2}{*}{Data amount} & \multirow{2}{*}{Initialization} & \multicolumn{2}{c|}{Overall} & \multicolumn{2}{c}{Car} & \multicolumn{2}{c}{Pedestrian} & \multicolumn{2}{c}{Cyclist} \\ \cmidrule(l){3-10} 
 &  & L2 mAP & L2 mAPH & L2 mAP & L2 mAPH & L2 mAP & L2 mAPH & L2 mAP & L2 mAPH \\ \midrule
\multirow{4}{*}{5\%} & Random & 42.59 & 38.83 & 50.09 & 49.59 & 53.88 & 44.06 & 23.79 & 22.85 \\
 & PointContrast\cite{PointContrast} & 44.48 & 40.55 & 51.87 & 51.37 & 55.36 & 45.03 & 26.22 & 25.24 \\
 & ProposalContrast\cite{proposalcontrast} & 45.21 & 41.45 & 52.29 & 51.82 & 56.23 & 46.28 & 27.10 & 26.24 \\
 & \textbf{MV-JAR (Ours)} & \textbf{47.93}\improve{5.34} & \textbf{44.50}\improve{5.67} & 56.22 & 55.78 & 58.80 & 49.77 & 28.75 & 27.95 \\ \midrule
\multirow{4}{*}{10\%} & Random & 54.85 & 51.22 & 54.95 & 54.51 & 62.11 & 52.76 & 47.49 & 46.40 \\
 & PointContrast\cite{PointContrast} & 54.80 & 51.02 & 55.41 & 54.95 & 60.56 & 50.86 & 48.44 & 47.24 \\
 & ProposalContrast\cite{proposalcontrast} & 54.77 & 51.09 & 55.64 & 55.20 & 60.54 & 51.16 & 48.14 & 46.92 \\
 & \textbf{MV-JAR (Ours)} & \textbf{58.29}\improve{3.44} & \textbf{54.99}\improve{3.77} & 59.17 & 58.74 & 64.58 & 56.02 & 51.12 & 50.20 \\ \bottomrule
\end{tabular}
}
\end{table*}
\section{Effects of Pre-training on Varying Distances}
In order to investigate the influence of pre-training across diverse distances, we present the L2 mAPH performance using 5\% fine-tuning data for different distance intervals in \cref{tab:performance_across_distance}. It can be observed that the performance enhancement of MVR declines as the distances increase, primarily boosting MVJ (i.e., MV-JAR) within the 0m-30m range. A plausible explanation for this phenomenon is the reduction in point density at greater distances, which 
\begin{table*}[h!]
\centering
\caption{Average results on the Waymo validation set, averaged across SST fine-tuned with Subset 0-2.}
\label{tab:average}
\scalebox{0.85}{\tablestyle{8pt}{1.0}
\begin{tabular}{@{}c|c|cc|cccccc@{}}
\toprule
\multirow{2}{*}{Data amount} & \multirow{2}{*}{Initialization} & \multicolumn{2}{c|}{Overall} & \multicolumn{2}{c}{Car} & \multicolumn{2}{c}{Pedestrian} & \multicolumn{2}{c}{Cyclist} \\ \cmidrule(l){3-10} 
 &  & L2 mAP & L2 mAPH & L2 mAP & L2 mAPH & L2 mAP & L2 mAPH & L2 mAP & L2 mAPH \\ \midrule
\multirow{4}{*}{5\%} & Random & 44.91 & 40.96 & 50.45 & 49.94 & 52.77 & 42.53 & 31.52 & 30.40 \\
 & PointContrast\cite{PointContrast} & 46.26 & 42.25 & 52.11 & 51.61 & 53.84 & 43.40 & 32.82 & 31.74 \\
 & ProposalContrast\cite{proposalcontrast} & 47.23 & 43.28 & 52.58 & 52.10 & 54.61 & 44.37 & 34.50 & 33.38 \\
 & \textbf{MV-JAR (Ours)} & \textbf{50.39}\improve{5.48} & \textbf{46.72}\improve{5.76} & 56.45 & 56.00 & 57.99 & 48.36 & 36.74 & 35.81 \\ \midrule
\multirow{4}{*}{10\%} & Random & 55.04 & 51.28 & 55.01 & 54.55 & 61.09 & 51.44 & 49.02 & 47.84 \\
 & PointContrast\cite{PointContrast} & 54.57 & 50.75 & 55.26 & 54.80 & 59.85 & 50.05 & 48.61 & 47.41 \\
 & ProposalContrast\cite{proposalcontrast} & 54.75 & 50.96 & 55.47 & 55.01 & 60.19 & 50.51 & 48.60 & 47.37 \\
 & \textbf{MV-JAR (Ours)} & \textbf{58.12}\improve{3.08} & \textbf{54.72}\improve{3.44} & 58.84 & 58.41 & 63.77 & 55.03 & 51.74 & 50.73 \\ \bottomrule
\end{tabular}
}
\end{table*}
makes the dense point clusters at closer ranges less ambiguous for the model to reconstruct voxel shapes. MVJ consistently outperforms MVR across various distance intervals, reinforcing our hypothesis that capturing voxel distributions plays a more crucial role in the model's representation learning. This is because LiDAR detectors downsample points into voxels to facilitate perception.

{\small
\bibliographystyle{ieee_fullname}
\bibliography{egbib}

\begin{thebibliography}{10}\itemsep=-1pt

\bibitem{beit}
Hangbo Bao, Li Dong, and Furu Wei.
\newblock Beit: Bert pre-training of image transformers.
\newblock {\em arXiv:2106.08254}, 2021.

\bibitem{gpt-3}
Tom Brown, Benjamin Mann, Nick Ryder, Melanie Subbiah, Jared~D Kaplan, Prafulla
  Dhariwal, Arvind Neelakantan, Pranav Shyam, Girish Sastry, Amanda Askell,
  Sandhini Agarwal, Ariel Herbert-Voss, Gretchen Krueger, Tom Henighan, Rewon
  Child, Aditya Ramesh, Daniel Ziegler, Jeffrey Wu, Clemens Winter, Chris
  Hesse, Mark Chen, Eric Sigler, Mateusz Litwin, Scott Gray, Benjamin Chess,
  Jack Clark, Christopher Berner, Sam McCandlish, Alec Radford, Ilya Sutskever,
  and Dario Amodei.
\newblock Language models are few-shot learners.
\newblock In {\em NeurIPS}, 2020.

\bibitem{nuscenes}
Holger Caesar, Varun Bankiti, Alex~H. Lang, Sourabh Vora, Venice~Erin Liong,
  Qiang Xu, Anush Krishnan, Yu Pan, Giancarlo Baldan, and Oscar Beijbom.
\newblock nuscenes: A multimodal dataset for autonomous driving.
\newblock {\em arXiv:1903.11027}, 2019.

\bibitem{swav}
Mathilde Caron, Ishan Misra, Julien Mairal, Priya Goyal, Piotr Bojanowski, and
  Armand Joulin.
\newblock Unsupervised learning of visual features by contrasting cluster
  assignments.
\newblock In {\em NeurIPS}, 2020.

\bibitem{co3}
Runjian Chen, Yao Mu, Runsen Xu, Wenqi Shao, Chenhan Jiang, Hang Xu, Yu Qiao,
  Zhenguo Li, and Ping Luo.
\newblock {CO}3: Cooperative unsupervised 3d representation learning for
  autonomous driving.
\newblock In {\em ICLR}, 2023.

\bibitem{simclr}
Ting Chen, Simon Kornblith, Mohammad Norouzi, and Geoffrey Hinton.
\newblock A simple framework for contrastive learning of visual
  representations.
\newblock In {\em ICML}, 2020.

\bibitem{cae}
Xiaokang Chen, Mingyu Ding, Xiaodi Wang, Ying Xin, Shentong Mo, Yunhao Wang,
  Shumin Han, Ping Luo, Gang Zeng, and Jingdong Wang.
\newblock Context autoencoder for self-supervised representation learning.
\newblock {\em arXiv:2202.03026}, 2022.

\bibitem{mocov2}
Xinlei Chen, Haoqi Fan, Ross Girshick, and Kaiming He.
\newblock Improved baselines with momentum contrastive learning.
\newblock {\em arXiv:2003.04297}, 2020.

\bibitem{mmdet3d2020}
MMDetection3D Contributors.
\newblock {MMDetection3D: OpenMMLab} next-generation platform for general {3D}
  object detection.
\newblock \url{https://github.com/open-mmlab/mmdetection3d}, 2020.

\bibitem{bert}
Jacob Devlin, Ming-Wei Chang, Kenton Lee, and Kristina Toutanova.
\newblock Bert: Pre-training of deep bidirectional transformers for language
  understanding.
\newblock {\em arXiv:1810.04805}, 2018.

\bibitem{vit}
Alexey Dosovitskiy, Lucas Beyer, Alexander Kolesnikov, Dirk Weissenborn,
  Xiaohua Zhai, Thomas Unterthiner, Mostafa Dehghani, Matthias Minderer, Georg
  Heigold, Sylvain Gelly, Jakob Uszkoreit, and Neil Houlsby.
\newblock An image is worth 16x16 words: Transformers for image recognition at
  scale.
\newblock In {\em ICLR}, 2021.

\bibitem{SST}
Lue Fan, Ziqi Pang, Tianyuan Zhang, Yu-Xiong Wang, Hang Zhao, Feng Wang, Naiyan
  Wang, and Zhaoxiang Zhang.
\newblock Embracing single stride 3d object detector with sparse transformer.
\newblock In {\em CVPR}, 2022.

\bibitem{Video-MAE}
Christoph Feichtenhofer, Haoqi Fan, Yanghao Li, and Kaiming He.
\newblock Masked autoencoders as spatiotemporal learners.
\newblock {\em arXiv:2205.09113}, 2022.

\bibitem{KITTI_Det}
Andreas Geiger, Philip Lenz, and Raquel Urtasun.
\newblock Are we ready for autonomous driving? the kitti vision benchmark
  suite.
\newblock In {\em CVPR}, 2012.

\bibitem{MAE}
Kaiming He, Xinlei Chen, Saining Xie, Yanghao Li, Piotr Doll{\'a}r, and Ross
  Girshick.
\newblock Masked autoencoders are scalable vision learners.
\newblock In {\em CVPR}, 2022.

\bibitem{moco}
Kaiming He, Haoqi Fan, Yuxin Wu, Saining Xie, and Ross Girshick.
\newblock Momentum contrast for unsupervised visual representation learning.
\newblock In {\em CVPR}, 2020.

\bibitem{scenecontrast}
Ji Hou, Benjamin Graham, Matthias Nie{\ss}ner, and Saining Xie.
\newblock Exploring data-efficient 3d scene understanding with contrastive
  scene contexts.
\newblock In {\em CVPR}, 2021.

\bibitem{strl}
Siyuan Huang, Yichen Xie, Song-Chun Zhu, and Yixin Zhu.
\newblock Spatio-temporal self-supervised representation learning for 3d point
  clouds.
\newblock In {\em CVPR}, 2021.

\bibitem{pointpillars}
Alex~H Lang, Sourabh Vora, Holger Caesar, Lubing Zhou, Jiong Yang, and Oscar
  Beijbom.
\newblock Pointpillars: Fast encoders for object detection from point clouds.
\newblock In {\em CVPR}, 2019.

\bibitem{gcc-3d}
Hanxue Liang, Chenhan Jiang, Dapeng Feng, Xin Chen, Hang Xu, Xiaodan Liang, Wei
  Zhang, Zhenguo Li, and Luc Van~Gool.
\newblock Exploring geometry-aware contrast and clustering harmonization for
  self-supervised 3d object detection.
\newblock In {\em CVPR}, 2021.

\bibitem{maskpoint}
Haotian Liu, Mu Cai, and Yong~Jae Lee.
\newblock Masked discrimination for self-supervised learning on point clouds.
\newblock In {\em ECCV}, 2022.

\bibitem{swin-transformer}
Ze Liu, Yutong Lin, Yue Cao, Han Hu, Yixuan Wei, Zheng Zhang, Stephen Lin, and
  Baining Guo.
\newblock Swin transformer: Hierarchical vision transformer using shifted
  windows.
\newblock In {\em CVPR}, 2021.

\bibitem{voxelformer}
Jiageng Mao, Yujing Xue, Minzhe Niu, Haoyue Bai, Jiashi Feng, Xiaodan Liang,
  Hang Xu, and Chunjing Xu.
\newblock Voxel transformer for 3d object detection.
\newblock In {\em CVPR}, 2021.

\bibitem{3DTR}
Ishan Misra, Rohit Girdhar, and Armand Joulin.
\newblock An end-to-end transformer model for 3d object detection.
\newblock In {\em CVPR}, 2021.

\bibitem{pointformer}
Xuran Pan, Zhuofan Xia, Shiji Song, Li~Erran Li, and Gao Huang.
\newblock 3d object detection with pointformer.
\newblock In {\em CVPR}, 2021.

\bibitem{point-mae}
Yatian Pang, Wenxiao Wang, Francis~EH Tay, Wei Liu, Yonghong Tian, and Li Yuan.
\newblock Masked autoencoders for point cloud self-supervised learning.
\newblock In {\em ECCV}, 2022.

\bibitem{pointnet_plus}
Charles~Ruizhongtai Qi, Li Yi, Hao Su, and Leonidas~J. Guibas.
\newblock Pointnet++: Deep hierarchical feature learning on point sets in a
  metric space.
\newblock In {\em NeurIPS}, 2017.

\bibitem{gpt-2}
Alec Radford, Jeffrey Wu, Rewon Child, David Luan, Dario Amodei, Ilya
  Sutskever, et~al.
\newblock Language models are unsupervised multitask learners.
\newblock {\em OpenAI blog}, 2019.

\bibitem{reconstructsapce}
Jonathan Sauder and Bjarne Sievers.
\newblock Self-supervised deep learning on point clouds by reconstructing
  space.
\newblock In {\em NeurIPS}, 2019.

\bibitem{Waymo}
Pei Sun, Henrik Kretzschmar, Xerxes Dotiwalla, Aurelien Chouard, Vijaysai
  Patnaik, Paul Tsui, James Guo, Yin Zhou, Yuning Chai, Benjamin Caine, Vijay
  Vasudevan, Wei Han, Jiquan Ngiam, Hang Zhao, Aleksei Timofeev, Scott
  Ettinger, Maxim Krivokon, Amy Gao, Aditya Joshi, Yu Zhang, Jonathon Shlens,
  Zhifeng Chen, and Dragomir Anguelov.
\newblock Scalability in {{Perception}} for {{Autonomous Driving}}: {{Waymo
  Open Dataset}}.
\newblock In {\em CVPR}, 2020.

\bibitem{SWFormer}
Pei Sun, Mingxing Tan, Weiyue Wang, Chenxi Liu, Fei Xia, Zhaoqi Leng, and
  Dragomir Anguelov.
\newblock Swformer: Sparse window transformer for 3d object detection in point
  clouds.
\newblock In {\em ECCV}, 2022.

\bibitem{Video-MAEW}
Zhan Tong, Yibing Song, Jue Wang, and Limin Wang.
\newblock Video{MAE}: Masked autoencoders are data-efficient learners for
  self-supervised video pre-training.
\newblock In {\em NeurIPS}, 2022.

\bibitem{reconfig_voxels}
Tai Wang, Xinge Zhu, and Dahua Lin.
\newblock Reconfigurable voxels: A new representation for lidar-based point
  clouds.
\newblock In {\em CoRL}, 2020.

\bibitem{balancedCD}
Tong Wu, Liang Pan, Junzhe Zhang, Tai Wang, Ziwei Liu, and Dahua Lin.
\newblock Balanced chamfer distance as a comprehensive metric for point cloud
  completion.
\newblock In {\em NeurIPS}, 2021.

\bibitem{PointContrast}
Saining Xie, Jiatao Gu, Demi Guo, Charles~R Qi, Leonidas Guibas, and Or Litany.
\newblock Pointcontrast: Unsupervised pre-training for 3d point cloud
  understanding.
\newblock In {\em ECCV}, 2020.

\bibitem{SimMiM}
Zhenda Xie, Zheng Zhang, Yue Cao, Yutong Lin, Jianmin Bao, Zhuliang Yao, Qi
  Dai, and Han Hu.
\newblock Simmim: A simple framework for masked image modeling.
\newblock In {\em CVPR}, 2022.

\bibitem{mae_for_audio}
Hu Xu, Juncheng Li, Alexei Baevski, Michael Auli, Wojciech Galuba, Florian
  Metze, Christoph Feichtenhofer, et~al.
\newblock Masked autoencoders that listen.
\newblock {\em arXiv:2207.06405}, 2022.

\bibitem{SECOND}
Yan Yan, Yuxing Mao, and Bo Li.
\newblock Second: Sparsely embedded convolutional detection.
\newblock {\em Sensors}, 2018.

\bibitem{foldingnet}
Yaoqing Yang, Chen Feng, Yiru Shen, and Dong Tian.
\newblock Foldingnet: Point cloud auto-encoder via deep grid deformation.
\newblock In {\em CVPR}, 2018.

\bibitem{proposalcontrast}
Junbo Yin, Dingfu Zhou, Liangjun Zhang, Jin Fang, Cheng-Zhong Xu, Jianbing
  Shen, and Wenguan Wang.
\newblock Proposalcontrast: Unsupervised pre-training for lidar-based 3d object
  detection.
\newblock In {\em ECCV}, 2022.

\bibitem{centerpoint}
Tianwei Yin, Xingyi Zhou, and Philipp Krahenbuhl.
\newblock Center-based 3d object detection and tracking.
\newblock In {\em CVPR}, 2021.

\bibitem{point-bert}
Xumin Yu, Lulu Tang, Yongming Rao, Tiejun Huang, Jie Zhou, and Jiwen Lu.
\newblock Point-bert: Pre-training 3d point cloud transformers with masked
  point modeling.
\newblock In {\em CVPR}, 2022.

\bibitem{position_prediction}
Shuangfei Zhai, Navdeep Jaitly, Jason Ramapuram, Dan Busbridge, Tatiana
  Likhomanenko, Joseph~Yitan Cheng, Walter Talbott, Chen Huang, Hanlin Goh, and
  Joshua Susskind.
\newblock Position prediction as an effective pretraining strategy.
\newblock In {\em ICML}, 2022.

\bibitem{point-m2ae}
Renrui Zhang, Ziyu Guo, Peng Gao, Rongyao Fang, Bin Zhao, Dong Wang, Yu Qiao,
  and Hongsheng Li.
\newblock Point-m2ae: multi-scale masked autoencoders for hierarchical point
  cloud pre-training.
\newblock {\em arXiv:2205.14401}, 2022.

\bibitem{zhang2022densesiam}
Wenwei Zhang, Jiangmiao Pang, Kai Chen, and Chen~Change Loy.
\newblock Dense siamese network for dense unsupervised learning.
\newblock In {\em ECCV}, 2022.

\bibitem{depthcontrast}
Zaiwei Zhang, Rohit Girdhar, Armand Joulin, and Ishan Misra.
\newblock Self-supervised pretraining of 3d features on any point-cloud.
\newblock In {\em CVPR}, 2021.

\bibitem{ibot}
Jinghao Zhou, Chen Wei, Huiyu Wang, Wei Shen, Cihang Xie, Alan Yuille, and Tao
  Kong.
\newblock ibot: Image bert pre-training with online tokenizer.
\newblock {\em arXiv:2111.07832}, 2021.

\bibitem{mae_for_medical}
Lei Zhou, Huidong Liu, Joseph Bae, Junjun He, Dimitris Samaras, and Prateek
  Prasanna.
\newblock Self pre-training with masked autoencoders for medical image
  analysis.
\newblock {\em arXiv:2203.05573}, 2022.

\bibitem{voxelnet}
Yin Zhou and Oncel Tuzel.
\newblock Voxelnet: End-to-end learning for point cloud based 3d object
  detection.
\newblock In {\em CVPR}, 2018.

\bibitem{ssn}
Xinge Zhu, Yuexin Ma, Tai Wang, Yan Xu, Jianping Shi, and Dahua Lin.
\newblock Ssn: Shape signature networks for multi-class object detection from
  point clouds.
\newblock In {\em ECCV}, 2020.

\end{thebibliography}
}

\end{document}